# Non-Linear Predictive Vector Quantization of speech

*Marcos Faúndez-Zanuy*

Escola Universitària Politècnica de Mataró
Universitat Politècnica de Catalunya (UPC)
Avda. Puig i Cadafalch 101-111, E-08303 Mataró (BARCELONA) SPAIN
`faundez@eupmt.es`

**Abstract**

In this paper we propose a Non-Linear Predictive Vector quantizer (PVQ) for speech coding, based on Multi-Layer Perceptrons. We also propose a method to evaluate if a quantizer is well designed, and if it exploits the correlation between consecutive outputs. Although the results of the Non-linear PVQ do not improve the results of the non-linear scalar predictor, we check that there is some room for the PVQ improvement.

## 1. Introduction

In [1] we proposed a scheme for nonlinear vectorial predictor based on neural nets. In this paper we apply the predictor for speech coding. This scheme is known as Non-Linear Predictive Vector Quantization [2, chap. 13] NL-PVQ. This system is similar to an ADPCM speech coder, where the NL predictor replaces the LPC predictor in order to obtain an ADPCM scheme with non-linear vectorial prediction. In addition, the scalar quantizer is replaced by a vectorial quantizer.

This scheme is a backward-adaptive ADPCM scheme. Thus, the coefficients are computed over the previous frame, and it is not needed to transmit the coefficients of the predictor, because the receiver has already decoded the previous frame and can obtain the same set of coefficients.

The optimization of a PVQ encoder needs a vectorial predictor and a vectorial quantizer. Sections 2 and 3 deal with these blocks. Section 4 proposes a method for the evaluation of a quantizer, and section 5 summarizes the main conclusions of this paper.

## 2. Vectorial nonlinear prediction

Our nonlinear predictor consists on a Multi Layer Perceptron (MLP) with 10 inputs, 2 neurons in the hidden layer, and $N$ outputs, where $N$ is the dimension of the vectors. In this paper we use $N=2$. The selected training algorithm is the Levenberg-Marquardt, that computes the approximate Hessian matrix, because it is faster and achieves better results than the classical backpropagation algorithm. We also apply a multi-start algorithm with five random initializations for each neural net. In [3] we studied several training schemes, and we concluded that the most suitable is the combination between Bayesian regularization and a committee of neural nets (each neural net is the result of training one random initialization).

In [1] we showed that the computation of a vectorial predictor based on a MLP is not critical. The vectorial prediction training procedure can be interpreted as a particular case of neural net training with output hints (see figure 1). Thus, the generalization of the scalar predictor to a vectorial prediction does not imply a great difference with respect to the scalar predictor.

Figure 1. Vectorial predictor based on a MLP

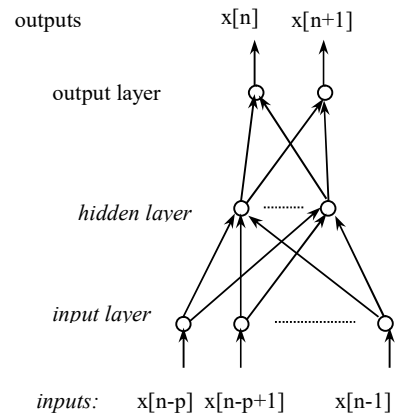

## 3. Vectorial quantizer

In order to design a vectorial quantizer (VQ) it is need a training sequence. The optimal design procedure must be iterative [4], because in a PVQ scheme the VQ is inside the loop of the ADPCM scheme. In order to achieve a "universal VQ", it should be obtained with as many speakers and sentences as possible and evaluated with a different database. In order to simplify the study (the nonlinear vectorial prediction computation is very time consuming using the MATLAB software), we have used only one speaker for VQ generation and 8 different speakers for PVQ system evaluation. We have used two different methods for codebook generation given a training sequence: random initialization plus the generalized Lloyd iteration, and the LBG algorithm [2]. We have used the following procedure:
1. A speech database is PVQ coded with a vectorial predictor and an adaptive scalar quantizer based on multipliers [5]. Although the prediction algorithm is vectorial, the residual error is scalar quantized, applying the scalar quantizer consecutively to each component of the residual vector.
2. We have used the residual signal of one sentence uttered by a female speaker (approximately 10000 vectors) and 3 quantization bits (Nq=3) as a training sequence. Figure 2 shows the visualization of the error sequences for several Nq.
3. A codebook is designed for several VQ sizes. Figures 3 and 4 show the obtained results with the random and the LBG method respectively.
4. The speech database is encoded and a new training sequence is computed for each VQ size. We use the respective residual errors of each VQ size (and speaker 1).
5. Go to step 3 and update the VQ using the training sequence of step 4 till a fixed number of iterations or a predefined criterion.

We have applied this closed-loop algorithm two times.

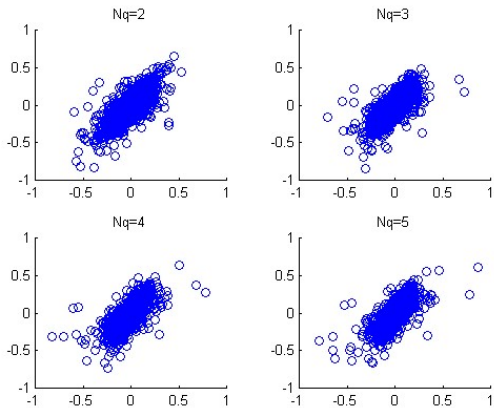

Figure 2. Initial training error sequence.

We believe that it is interesting the evaluation of both algorithms (random and LBG) because a very well fitted codebook to a given training sequence can be less robust when dealing with different vectors not used for training.

Figure 5 compares the quantization distortion of the initial training sequence using the VQ obtained with the random method and the LBG one. It is clear that the latest one is much better than the former (the quantization distortion is smaller). On the other hand, the obtained SEGSNR encoding the 8 speakers database does not show significative differences.

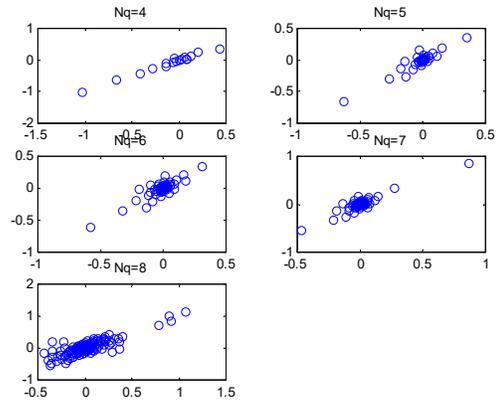

Figure 3. Codebooks with the random method.

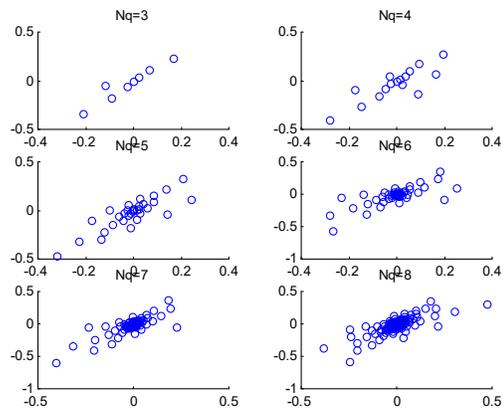

Figure 4. Codebooks with the LBG algorithm

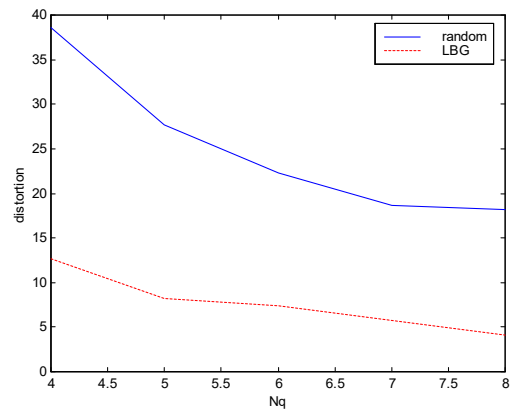

Figure 5. Distortion quantization with random and LBG VQ.

Table 1 compares the results with the classical scalar ADPCM, and the PVQ with scalar prediction. We have used the following notation:
- Q (Quantizer): scalar (scal) or vectorial quantizer (VQ) (scalar or VQ)

- PRED (Predictor): scalar (scal) or vectorial (rand= VQ with random algorithm and generalized lloyd iteration, LBG= VQ with the LBG algorithm).
- Nq: Equivalent number of quantization bits per sample. For instance, if the codebook size is 5 and the vector dimension is 2, then Nq=2.5

Table 1 SEGSNR for several schemes.

| Q | PRED | Nq=2 | | Nq=2.5 | | Nq=3 | | Nq=3.5 | | Nq=4 | |
|---|---|---|---|---|---|---|---|---|---|---|---|
| | | SEG SNR | σ | SEG SNR | σ | SEG SNR | σ | SEG SNR | σ | SEG SNR | σ |
| scal | scal | 14 | 5.6 | | | 20.8 | 6 | | | 26 | 6.6 |
| scal | vect | 12.9 | 4.8 | | | 18.8 | 5.3 | | | 23.6 | 5.5 |
| rand | vect | 10.4 | 8.3 | 15.8 | 6.9 | 19 | 5.9 | 21.4 | 6.1 | 25 | 5.7 |
| LBG | vect | 11.2 | 9.3 | 14.9 | 8 | 19.2 | 5.5 | 21.9 | 5.6 | 24.9 | 6 |

Table 2 shows the results after the iterative loop described in section 3. The results for the random VQ have been obtained using for each codebook size his respective error sequence, while the results for the LBG VQ have been obtained using the same error sequence for all the bit range.

Table 2 SEGSNR for several schemes, after one iteration.

| VQ | PRED | Nq=2 | | Nq=2.5 | | Nq=3 | | Nq=3.5 | | Nq=4 | |
|---|---|---|---|---|---|---|---|---|---|---|---|
| | | SEG SNR | σ | SEG SNR | σ | SEG SNR | σ | SEG SNR | σ | SEG SNR | σ |
| rand | vect | 12.1 | 7.9 | 15.4 | 7.5 | 19.3 | 6.3 | 21.1 | 6.2 | 25.4 | 6.3 |
| LBG | vect | 12.4 | 7.6 | 16.1 | 5.9 | 18.4 | 6.2 | 21.3 | 6.1 | 23.8 | 6.6 |

Comparing tables 1 and 2 it can be seen that there is little improvement on the SEGSNR. On the other hand, the PVQ system does not improve the results of the scalar one. In [1] we showed that the vectorial predictor offers good results, so we propose to make a deepest study of the vectorial quantizer in order to evaluate if this block is good enough or if it can be improved.

## 4. Study of the quantizer

The results of the previous section reveal that the classical ADPCM scheme with scalar prediction and quantization yields better results than the vectorial scheme. This is because the scalar quantizer is well fit to the linear predictor. In addition, the quantizer based on multipliers is a memory quantizer. That is: the quantization step depends on the previous sample. On the other hand, the VQ is a memoryless quantizer. In order to study the relevance of one sample memory quantization, we propose to evaluate the zero order entropy $H_0(X)$ and the first order entropy $H_1(X)$ of the codewords, where:

- $H_0(x) = \sum_{i=1}^{M} P_i \log_2 \frac{1}{P_i}$

- $H_1(X) = \sum_{j=1}^{M}\sum_{i=1}^{M} P(ij) \log_2 \frac{1}{P(i|j)}$

- $P_i$ is the probability of the codeword $i$.
- $P(i|j)$ is the probability of the codeword $i$ knowing that the previous codeword has been the codeword $j$.

It is important to take into account that this formulation is valid for scalar and vectorial quantization. The unique difference is that in the former case each codeword is equivalent to one sample, while in the latest one each codeword is equivalent to a vector (group of samples).

It would be interesting to study higher order entropies, but the amount of required data and the computational burden makes this evaluation unpractical.

The better designed the quantizer, the higher the entropy, because all the codewords have the same probability of being chosen. In this case, $H_0(X) \cong N_q$.

Otherwise, the outputs of the quantizer (codewords) can be encoded with a lossless method (for example Huffmann) in order to reduce the data rate.

On the other hand, if $H_1(X) << H_0(X)$ means that there is a strong correlation between consecutive quantizer outputs, and two observations can be made:

1. The outputs of the quantizer (codewords) can be encoded with a lossless method (for example Huffmann) in order to reduce the data rate.
2. The quantizer can be improved taking into account the previous sample (using a memory quantizer). The goal is to obtain $H_1(X) \cong H_0(X) \cong N_q$ (remember that $H_1(X) \leq H_0(X) \leq N_q$ by definition). In this case, all the codewords are equal probably used, and $P(x[n]|x[n-1]) \cong P(x[n])$, so the quantizer has removed the first order dependencies, and no improvement is achieved by using a Huffmann code.

Our goal is the latest observation, rather than the former one, because the better the quantizer, the better the prediction. If the entropy is smaller than $N_q$ means that some codewords are not used, so the useful number of quantization bits is smaller than $N_q$.

Table 3 shows the zero-order and first-order entropies of the quantization codewords, obtained with scalar linear prediction and scalar quantization based on multipliers. An important fact when computing high order entropies is the size of the database. In order to check if the database is big enough, these values have been obtained with two different databases:
a) One sentence of one speaker (23800 samples).
b) Eight sentences of eight different speakers (168000 samples).

Comparing the obtained results in both situations it can be check that one sentence is enough, because the results are very similar increasing eight times de database size. For this reason, the next experiment will be computed using only one sentence.

Table 3 Entropies for scalar quantization and scalar prediction.

|  | $N_q=2$ | | $N_q=3$ | | $N_q=4$ | | $N_q=5$ | |
|---|---|---|---|---|---|---|---|---|
| speakers | $H_0$ | $H_1$ | $H_0$ | $H_1$ | $H_0$ | $H_1$ | $H_0$ | $H_1$ |
| 1 | 1.90 | 1.87 | 2.73 | 2.7 | 3.6 | 3.56 | 4.39 | 4.35 |
| 8 | 1.90 | 1.87 | 2.72 | 2.7 | 3.58 | 3.55 | 4.37 | 4.33 |

Table 4 shows the entropies using vector quantization and vectorial prediction. The codebook size ranges from 4 to 8 bits. In order to facilitate the comparison with table 3, it is represented the equivalent entropy and number of quantization bit for each sample of the original speech signal (that is, the results have been divided by 2, obtaining an equivalent number of bits per sample ranging from 2 to 4 with a 0.5 step increment). The results have been obtained with two different algorithms for codebook generation: random method with generalized Lloyd iteration, and LBG algorithm.

Table 4 Entropies for VQ and vectorial prediction.

| Q | $N_q=2$ | | $N_q=2.5$ | | $N_q=3$ | | $N_q=3.5$ | | $N_q=4$ | |
|---|---|---|---|---|---|---|---|---|---|---|
|  | $H_0$ | $H_1$ | $H_0$ | $H_1$ | $H_0$ | $H_1$ | $H_0$ | $H_1$ | $H_0$ | $H_1$ |
| rand | 1.2 | 1 | 1.8 | 1.4 | 2.6 | 2.1 | 3.4 | 2.6 | 3.7 | 2.6 |
| LBG | 1.3 | 1.1 | 2.3 | 1.9 | 2.9 | 2.4 | 3.4 | 2.7 | 3.9 | 2.5 |

We have observed that the scalar-multipliers quantizer is able to remove the dependency of the previous sample (even for Nq=5, the results are $H_0$=4.37 and $H_1$=4.33). On the other hand, this is not valid for the VQ-based scheme.

We can observe that the VQ is better designed using the LBG algorithm (all he codewords have the same chance, yielding $H_0 \cong N_q$), but it does not exploit the redundancy between consecutive codewords, because the first order entropy is significantly smaller than the zero order one. Thus, there is room for VQ improvement, probably using a finite state VQ (in this scheme the output of the quantizer depends on the previous output).

## 5. Conclusion

In this paper we have proposed a Non-linear Predictive Vector Quantization speech encoder based on a multilayer perceptron. We think that this scheme could be a preliminar step towards a more sophisticated coder like the CELP scheme.

We have also proposed a methodology for the analysis of a quantizer, based on two propositions:
1. If the quantizer is well designed, the zero order entropy will be approximately equal to the number of quantization bits (that is, all the quantization outputs have the same probability).
2. If the quantizer exploits the correlation between consecutive outputs, the first order entropy will be approximately the same than the zero order entropy. Otherwise, it will be smaller.

Although we have not improved the results of the scalar prediction and quantization scheme we have analyzed the scalar and the vectorial quantizer, and we have obtained that there is quite room for VQ improvement but not for the scalar quantizer. We think that the scalar scheme outperforms the vectorial scheme because it uses a memory quantizer, while the vectorial scheme does not exploit the redundancy between consecutive quantizer outputs.

## 6. References

[1] M. Faúndez "Nonlinear prediction with neural net". Submitted to International Workshop on Artificial and Natural Neural Networks '2001, Granada
[2] A. Gersho & R. M. Gray, "Vectorial Quantization and signal compression". Ed. Kluwer 1992.
[3] M. Faúndez-Zanuy, "Nonlinear predictive models computation in ADPCM schemes". Vol. II, pp 813-816. EUSIPCO 2000, Tampere.
[4] V. Cuperman & A. Gersho, "Vector Predictive coding of speech at 16 kbits/s". IEEE Trans. on Comm. vol. COM-33, pp.685-696, July 1985.
[5] N. S. Jayant & P. Noll "digital compression of waveforms". Ed. Prentice Hall 1984.

### Acknowledgement

This work has been supported by the Spanish grant CICYT TIC2000-1669-C04-02